\documentclass[sn-mathphys]{sn-jnl}% Math and Physical Sciences Reference Style

\jyear{2021}%

\theoremstyle{thmstyleone}%
\usepackage{bm}
\usepackage{dsfont}

\theoremstyle{thmstyletwo}%

\theoremstyle{thmstylethree}%

\newcommand{\Break}{\State \textbf{break}}

\DeclareMathOperator*{\argmin}{arg\,min}

\begin{document}

\title[Using Sequential Statistical Tests for Efficient Hyperparameter Tuning]{Using Sequential Statistical Tests for Efficient Hyperparameter Tuning}


\author[1]{\fnm{Philip} \sur{Buczak}}\email{buczak@statistik.tu-dortmund.de}

\author[1]{\fnm{Andreas} \sur{Groll}}\email{groll@statistik.tu-dortmund.de}

\author[1,2]{\fnm{Markus} \sur{Pauly}}\email{pauly@statistik.tu-dortmund.de}

\author[3]{\fnm{Jakob} \sur{Rehof}}\email{jakob.rehof@tu-dortmund.de}

\author[1,2]{\fnm{Daniel} \sur{Horn}}\email{dhorn@statistik.tu-dortmund.de}

\affil[1]{\orgdiv{Department of Statistics}, \orgname{TU Dortmund University}, \orgaddress{%\street{Vogelpothsweg 87}, 
\city{Dortmund}, \postcode{44227}, \country{Germany}}}

\affil[2]{\orgdiv{Research Center Trustworthy Data Science and Security}, \orgname{UA Ruhr}, \orgaddress{%\street{Street},
\city{Dortmund}, \postcode{44227}, \country{Germany}}}

\affil[3]{\orgdiv{Department of Computer Science}, \orgname{TU Dortmund University}, \orgaddress{%\street{Vogelpothsweg 87}, 
\city{Dortmund}, \postcode{44227}, \country{Germany}}}

%\affil[3]{\orgdiv{Department}, \orgname{Organization}, \orgaddress{\street{Street}, \city{City}, \postcode{610101}, \state{State}, \country{Country}}}

\abstract{Hyperparameter tuning is one of the the most time-consuming parts in machine learning. Despite the existence of modern optimization algorithms that minimize the number of evaluations needed, evaluations of a single setting may still be expensive. Usually a resampling technique is used, where the machine learning method has to be fitted a fixed number of $k$ times on different training datasets. The respective mean performance of the $k$ fits is then used as performance estimator. 
Many hyperparameter settings could be discarded after less than $k$ resampling iterations if they are clearly inferior to high-performing settings. However, resampling is often performed until the very end, wasting a lot of computational effort. To this end, we propose the Sequential Random Search (SQRS) which extends the regular random search algorithm by a sequential testing procedure aimed at detecting and eliminating inferior parameter configurations early. We compared our SQRS with regular random search using multiple publicly available regression and classification datasets. Our simulation study showed that the SQRS is able to find similarly well-performing parameter settings while requiring noticeably fewer evaluations. Our results underscore the potential for integrating sequential tests into hyperparameter tuning.}

\keywords{Machine Learning, Hyperparameter Tuning, Sequential Testing}

\maketitle

\section{Introduction}\label{sec1}

Whether for sales prediction, predictive maintenance, sports forecasting, treatment recommendation, neuroimaging analysis or creativity research, machine learning (ML) models are widely used \cite{adewumi2017survey,bohanec2017explaining,huang2020travel,groll2019hybrid, susto2014adaptive, hahn2022, buczak2022}. Just as there is a plethora of application problems, there is an ever growing variety of ML algorithms aiming to provide best possible solutions. Thus, the main issue of applying ML often is identifying the algorithm which performs best at the task at hand. %However, as the famous \textit{no free lunch theorem} \cite{wolpert} predicates, no single algorithm performs best at predicting all tasks. 
The fact that most ML methods have a set of meta parameters (also called \textit{hyperparameters}) whose optimal choice is problem-specific aggravates the problem of algorithm selection. 

Usually, problem-optimal choices for hyperparameters are derived from a hyperparameter tuning process aimed at finding parameter settings that minimize the generalization error, i.e., the expected loss on unknown data from the same data generating process (DGP). However, the true generalization error is unknown and can only be estimated, e.g., using resampling methods such as $k$-fold cross-validation or bootstrapping. Thus, minimizing the generalization error is restricted to minimizing an (unbiased) estimate of it. In theory, this poses a stochastic optimization problem, which in practice is commonly approached through heuristically comparing a set of candidate settings from a pre-specified parameter search space. These settings can either be generated by \textit{grid} or \textit{random search} \cite{bergstra}. The candidate configuration which minimizes the resampling error then results as the optimal setting. Within the context of stochastic optimization, the resampling error is equivalent to the stochastic target function and the resampling strategy ultimately describes a repeated evaluation in the same parameter.         

A key advantage of the random search algorithm is the possibility of paral\-lelizing model evaluations. However, single evaluations within the resampling can still lead to high computational efforts depending on the learner and the dimensionality of the dataset. As such, it would be desirable to stop the evaluation process for parameter settings whose inferior quality is already apparent after a few evaluation steps. An early stopping could prevent redundant computations and potentially save a lot of run time.

The idea of early stopping is also at the core of statistical sequential test theory.
Contrary to regular statistical testing with a fixed sample size $n$, sequential tests dictate a process in which the decision to reject or accept the null hypothesis, or to continue sampling is determined at each sampling step anew. Therefore, sequential tests are especially useful when sampling is costly and it desirable to form a decision based on as few observations as possible. In addition, sequential tests control both, type I and II error, thus allowing for equal treatment of $H_0$ and $H_1$, whereas regular statistical tests only allow for rejecting $H_0$ \cite{siegmund}.

The aim of our work is to investigate the feasibility of employing sequential testing during hyperparameter tuning to save evaluations and computational time. In particular, we aim to answer the following two research questions: 
\begin{enumerate}
    \item Can a proper sequential test be constructed for use in the context of hyperparameter tuning?
    \item How does such an approach perform in comparison to a regular random search?
\end{enumerate}
We see studying these two questions as a crucial first step that may open many further avenues of making the hyperparameter tuning process more efficient.

In the \hyperref[sec:ml]{next} section, we will give a brief introduction to the general ML process and how hyperparameter tuning is incorporated. We introduce the datasets and ML algorithms used in our work in Section~\hyperref[sec:dat]{3}. In Section~\hyperref[sec:test]{4}, we  determine a suitable sequential test and use it to extend the regular random search in Section~\hyperref[sec:sqrs]{5}. We compare our algorithm and the regular random search in a simulation study on multiple datasets in Section~\hyperref[sec:sim]{6}. Finally, we review and discuss our findings in Section~\hyperref[sec:dis]{7}. 

\section{Machine Learning and Hyperparameter Tuning}
\label{sec:ml}
%\subsection{Machine Learning}
A basic object of ML is modeling a functional mapping $f:\mathcal{X} \rightarrow \mathcal{Y}$ between a vector of $p$ features $\bm X = (X_1, \dots, X_p)^T$ %which follows an (unknown) distribution $\mathcal{G}_{\bm X}$,
and a target variable $Y$. % with (also unknown) distribution $\mathcal{G}_Y$.
Since the true $f$ is usually unknown, a ML method is used to determine an approximation $\hat{f}$ that describes $f$ as well as possible. In the case of supervised learning this is done based on an annotated dataset consisting of $n$ pairs of observations of the form $\left(\bm x_i, y_i \right)_{i = 1, \dots, n}$, where $\bm x_i$ are the feature values of the $i$th observation and $y_i$ is the corresponding (true) target variable value. The goodness of $\hat{f}$ is assessed using a loss function $L(y, \hat{f}(\bm x))$. In the regression case, the squared (also called Gaussian) loss function is a common choice while the 0-1 loss function is commonly used in the classification context \cite{hastie}. 
% In the regression case, the squared (also called Gaussian) loss function 
% \[L_{\text{sq}}(y, \hat{f}(\bm x) ) = (y - \hat{f}(\bm x))^2\]
% is a common choice while the 0-1 loss function
% \[L_{\text{0-1}}(y, \hat{f}(\bm x)) = \mathds{1}_{y \neq \hat{f}(\bm x)} = 
% \begin{cases} 0, & y = \hat{f}(\bm x)\\ 1, & y \neq \hat{f}(\bm x) \end{cases}\] is commonly used in the classification context \cite{hastie}. 
The principal goal is to determine $\hat{f}$ such that it minimizes the generalization error (i.e., the expected loss over all possible data samples). %, i.e.\ \[
%\hat{f} = \argmin_{\tilde{f}} E_{\bm X, Y}\left( L\left(Y, \tilde{f}(\bm X) \right) \right).
%\] 
However, since the distributions of $\bm X$ and $Y$ are usually unknown and only finitely many data points are available, $\hat{f}$ is instead obtained by optimizing the estimated generalization error. Because one is generally interested in predicting unknown data as best as possible, the original dataset is usually split into a training and test set. The training set is then used for fitting the model, while the test set is used for evaluating the performance of the model, i.e. for estimating generalization error. Theoretically, it would be possible to use the same data to train and test the model. However, such an approach is problematic because the model is determined to explain the training data as well as possible. Thus, learning and validating on the same data leads to biased, too optimistic error rates (overfitting;~\cite{hastie}). There exists a variety of different resampling techniques to generate training and test sets from an original dataset. Commonly used techniques include $k$-fold cross-validation~\cite{hastie} and bootstrapping~\cite{bootstrap} .

%on a test dataset $D^{\text{test}}$, consisting of pairs of observations $\left(\bm x^{\text{test}}_i, y^{\text{test}}_i \right)_{i = 1, \dots, n^{\text{test}}}$. %, i.e.
% \[
% \hat{f} = \argmin_{\tilde{f}} \frac{1}{n^{\text{test}}}\sum^{n^{\text{test}}}_{i=1} L\left(y^{\text{test}}_i, \tilde{f}(\bm x^{\text{test}}_i) \right).
% \] 
When using the squared and 0-1 loss functions, the generalization error is estimated on the test set consisting of $n^{\text{test}}$ observation pairs $\left(\bm x^{\text{test}}_i, y^{\text{test}}_i \right)_{i = 1, \dots, n^{\text{test}}}$ via the \textit{Mean Squared Error} (MSE) and \textit{Mean Misclassification Error} (MMCE) performance measures, respectively, where
\[\text{MSE} = \frac{1}{n^{\text{test}}}\sum^{n^{\text{test}}}_{i=1} \left( y^{\text{test}}_i - \hat{f}(\bm x^{\text{test}}_i) \right)^2 \ \text{and} \ \text{MMCE} = \frac{1}{n^{\text{test}}}\sum^{n^{\text{test}}}_{i=1} \mathds{1}_{y^{\text{test}}_i \neq \hat{f}(\bm x^{\text{test}}_i)}.\]

%Figure \ref{cycle} schematically represents the flow of a machine learning process described above.
% Figure \ref{cycle} displays the flow of a machine learning experiment as described above. 
%Theoretically, it would be possible to use the same data to learn and to validate the model, i.e.\ $D^{\text{learn}} = D^{\text{test}}$. However, such an approach is problematic because the model is determined to explain the training data as well as possible. Thus, learning and validating on the same data leads to biased, too optimistic error rates (overfitting; \cite{hastie}). In general, it is desirable to predict unknown observations as well as possible with the model. Therefore, only observations that were not already part of the training data should be used for model validation. %A variety of different resampling procedures for the construction of test data exists, like cross-validation or bootstrapping.

An additional challenge in machine learning  % that often arises in determining an optimal model $\hat{f}$ 
is that many learning methods have additional hyperparameters $\bm \lambda \in \Lambda$ besides the internally determined model parameters, which must be specified a priori and whose optimality is problem-specific.
Thus, for a fixed $\hat{f}_{\bm \lambda}$ the original optimization problem is expanded by an outer optimization problem, which as before can only be approximated by
\[
\bm \lambda^\star = \argmin_{\lambda \in \Lambda} \frac{1}{n^{\text{test}}}\sum^{n^{\text{test}}}_{i=1} L\left(y^{\text{test}}_i, \hat{f}_\lambda(\bm x^{\text{test}}_i) \right).
\] 

This optimization problem is also called {\it hyperparameter tuning problem}. Many algorithms have been developed to solve this problem. All such tuning algorithms work in a similar way: a set $\tilde \Lambda \subset \Lambda$ of multiple hyperparameter settings is proposed, the respective values of the loss function are estimated using a resampling procedure and finally, the value $\lambda \in \tilde \Lambda$ leading to the smallest loss is used. The strategy of choosing the set $\tilde \Lambda$, however, differs for most optimization algorithms. While random and grid search operate on a single large batch, population-based methods such as evolutionary algorithms (EA) keep a population of parameter configurations that is continually optimized. This is achieved by recombining or mutating (i.e.\ locally transforming) already existing configurations into new candidates or sampling from them from distributions updated via the current population members. A popular EA commonly used for general black-box optimization problems is the \textit{Covariance Matrix Adaptation Evolution Strategy} (CMA-ES, \cite{cmaes}) which has also been adapted specifically for the hyperparameter tuning of SVMs \cite{svmCMAES} and neural networks \cite{nnCMAES}, for example. For further uses of EAs in hyperparameter tuning, see e.g., \cite{bochinski} and \cite{young}.

%Apart from the model agnostic approaches described so far, 
The class of \textit{Sequential Model-Based Optimization} (SMBO, also known as Bayesian Optimization) uses two components for optimizing parameter configurations: a probabilistic surrogate model and an acquisition function that is usually cheap to evaluate. The surrogate model is updated iteratively based on previous evaluations, while the acquisition function determines suitable new candidates for evaluation. A popular choice are Gaussian processes for the surrogate model combined with the \textit{Expected Improvement} as acquisition function \cite{feurer}. As alternative to Gaussian processes, neural networks \cite{smboNN}, random forests \cite{smboRF} and tree parzen estimators \cite{bergstraAlg} are used as well. Similar to SMBO, Monte-Carlo Tree Search (MCTS;~\cite{kocsis2006}) has also been applied to hyperparameter tuning \cite{rakoto2019}. MCTS combines the classic tree search with ideas from Reinforcement Learning, exploring its search space and iteratively focusing on the most promising regions.

Another commonly employed approach in hyperparameter tuning is the reduction of evaluations, e.g., by eliminating suboptimal parameter configurations. An early example of this are the \textit{Hoeffding Races} \cite{hoeffdingrace}, in which bounds (resembling a confidence interval) are placed around the error estimates and iteratively updated. If the lower error bound of one model is greater than the upper error bound of at least another model, the former model is discarded. The concept of racing has been modified and extended, for example in \cite{domingos} or \cite{mnih}. Another popular variation is the \textit{F-Race} algorithm \cite{frace}, which eliminates bad parameter configurations via the Friedman test. The \textit{Iterated F-Race} algorithm \cite{itFrace} and its extension, \textit{Iterated Racing} \cite{irace}, added a population component to the original F-Race algorithm. Similar to us, Krueger~et~al.~\cite{krueger} also employ a sequential test for detecting and eliminating weak parameter configurations early. However, their sequential test procedure does not operate directly on the error estimates but on indicators derived from them. Further details on existing hyperparameter optimization algorithms can be found in the review papers \cite{bergstra2011} and \cite{yu2020review}.

\section{Datasets and Machine Learning Algorithms}
\label{sec:dat}
Throughout this work, we benchmarked the performance of multiple ML methods on five regression and five binary classification datasets, see Table~\ref{dataSets}. %provides an overview of the datasets we use in our simulation studies. 
The dataset \textit{Insurance} was taken from \cite{insurance}, \textit{Diamond} from \cite{ggplot}, and \textit{Wage} from \cite{ISLR}. The remaining datasets were obtained from the OpenML platform  \cite{openML}. The original \textit{Diamond} dataset contains 54\,000 observations. In our analysis, we used a random sample of 5\% of the original data.

\begin{table}[h!]
\begin{center}
\begin{minipage}{\textwidth}
\caption{Datasets used for simulation studies.}
\label{dataSets}
\setlength{\tabcolsep}{1.1mm}
\begin{tabular}{llccl}
\toprule
Type & Dataset & Obs.  &  Feat. & Description\\
\midrule
\multirow{11}{*}{Regression} & \multirow{2}{*}{\textit{Boston}}  & \multirow{2}{*}{506} & \multirow{2}{*}{13} & Median housing prices in the Boston area\\
& & & & based on neighborhood characteristics\\
\cmidrule{2-5}
& \textit{Insurance}  & 1\,338  & 6 & Medical insurance charges based on patients\\\cmidrule{2-5}
&\textit{Diamond}  & 2\,700 & 9 & Diamond prices based on cut characteristics\\
\cmidrule{2-5}
&\textit{Wage} & 3\,000  &  8 &  Wages based on socio-economic information\\
\cmidrule{2-5}
&\textit{Concrete} &  1\,030 & 8 & Concrete compressive strength based on\\
& & & & ingredients\\
\midrule
\multirow{13}{*}{Classification} & \textit{German Credit}  & 1\,000 & 20 & Credit risk based on customer attributes\\
\cmidrule{2-5}
%& & & & based on neighborhood characteristics\\
& \multirow{2}{*}{\textit{Phoneme}}  & \multirow{2}{*}{5\,404}  & \multirow{2}{*}{5} & Distinction of nasal and oral sounds based \\
& & & &  on frequency characteristics\\
\cmidrule{2-5}
&\multirow{2}{*}{\textit{Pima Indians}}  & \multirow{2}{*}{768} & \multirow{2}{*}{8} & Diabetes status in indigenous population\\
& & & &  based on diagnostic characteristics\\
\cmidrule{2-5}
&\textit{Cancer} & 569  &  30 &  Cancer recognition based on characteristics\\
& & & &of a fine needle aspirate of a breast mass\\
\cmidrule{2-5}
&\textit{Ionosphere} &  351 & 34 & Distinction of 'bad' and 'good' radar returns\\
& & & & in the ionosphere w.r.t. to free electrons\\
\bottomrule
\end{tabular}
\end{minipage}    
\end{center}
\end{table}

As ML algorithms, we used decision trees from the \texttt{R}-package \texttt{rpart} \cite{rpart}, random forest from the \texttt{ranger} package \cite{ranger}, XGBoost from the \texttt{xgboost} package  \cite{xgboost}, as well as elastic net linear regression from the \texttt{glmnet} package \cite{glmnet}. %\MP{Brauchen wir ein Argument warum wir uns auf diese fokussieren?}
Table~\ref{lrnTab} lists the considered hyperparameters and corresponding search spaces for each learning method.  Here, for decision trees (\texttt{rpart}), \texttt{cp} denotes the complexity parameter which specifies by how much a split must contribute to the improvement of the fit so that the corresponding sub-tree is not pruned. Moreover, the parameter \texttt{maxdepth} denotes the maximum tree depth. % of the tree. 
For the random forest, \texttt{mtry} describes the number of variables randomly drawn as split candidates, and \texttt{sample.fraction} and \texttt{replace} describe the fraction of observations used for each tree model and whether or not they are drawn with replacement.
For XGBoost, \texttt{nrounds} denotes the maximum number of iterations, \texttt{eta} the learning rate, and \texttt{max\textunderscore depth} the maximum depth of the trees used.
For the elastic net linear regression, \texttt{alpha} regulates the mixture parameter of the elastic net regularization and \texttt{lambda} the degree of penalization.
For the remaining hyperparameters of the individual methods, which are not subject of the optimization, the respective default settings were used.

\begin{center}
\begin{table}[h!]
\caption{Hyperparameters and search spaces used for tuning experiments.}
\label{lrnTab}
\begin{tabular}{llll}
\toprule
Method & \texttt{R} package & Hyperparameter & Search space\\
\midrule
\multirow{2}{*}{Decision Tree} & \multirow{2}{*}{\texttt{rpart}} & \texttt{cp} & $[0, 0.5]$ \\
& & \texttt{maxdepth} & $\{1, \dots, 30\}$ \\ 
\midrule
&  & \texttt{mtry} & $\{1, \dots, \#\text{Features}\}$\\
Random Forest &\texttt{ranger} & \texttt{replace} & \{\texttt{TRUE}, \texttt{FALSE}\}\\
& & \texttt{sample.fraction} & $[0.5, 1]$\\
\midrule
&  & \texttt{nrounds} & $\{2, \dots, 100\}$ \\
XGBoost &\texttt{xgboost} & \texttt{eta} & $[0.01, 1]$ \\
& & \texttt{max\textunderscore depth} & $\{1, \dots, 15\}$ \\
\midrule
\multirow{2}{*}{Elastic Net}&\multirow{2}{*}{\texttt{glmnet}} & \texttt{lambda} & $2^x \ \text{with} \ x \in [-15,15]$  \\
& & \texttt{alpha} & $[0, 1]$ \\
\midrule
\end{tabular}
\end{table} 
\end{center}

\section{Determining an Appropriate Sequential Test}
\label{sec:test}
%\subsection{Determining an Appropriate Sequential Test}
The main goal of our modified random search is to reduce the number of required evaluation steps while obtaining high performing solutions. As computing resampling errors can be viewed as a sequential process, we regard sequential statistical tests as a natural fit for this kind of situation. For a general overview on sequential statistical tests see, e.g., \cite{ghosh,siegmund}. An essential class of sequential tests are \textit{Sequential Probability Ratio Tests} (SPRT; \cite{wald}). 
For real parameters $\theta_0 < \theta_1$, the classic form of a SPRT for 
$H_0: \theta = \theta_0 \ \text{vs.} \ H_1: \theta = \theta_1 \ \text{with} \ \theta_0 < \theta_1$ can be described as follows: having sampled observation $u_n$ in step $n = 1, 2, \dots,$ calculate the test statistic 
\[Z_n := \ln \frac{f_n\left( u_1, \dots, u_n; \theta_1 \right)}{f_n\left( u_1, \dots, u_n; \theta_0 \right)}\,,\]
where $f_n\left(\cdot \ ; \theta_i\right)$ denotes the density function corresponding to $\theta_i, i = 0,1$. Then, 
\begin{itemize}
    \item[(i)] if $Z_n < b$, terminate and accept $H_0$,
    \item[(ii)] if $Z_n > a$, terminate and accept $H_1$,
    \item[(iii)] if $b < Z_n < a$, continue and sample a new observation $u_{n+1},$ 
\end{itemize}
where $b,a \in \mathbb{R}, \ b < a$ define the continuation region $[b,a]$ of the test and are chosen such that the type I and II errors are controlled for pre-specified $\alpha$ and $\beta$ values.   
%\red{[Mir kam es etwas ungewohnt vor, dass $b < a$, aber %vielleicht ist das auch in der Literatur immer so]}
% Ist tatsaechlich so :D

Many sequential tests are based on parametric assumptions. However, in the context of hyperparameter tuning it is not evident what kind of parametric properties can be assumed for the error estimates. As such, this necessitated an analysis of resampling error distributions beforehand to derive at least approximate parametric properties. Thus, we first performed a small simulation study in which we fit several common distribution families via standard maximum-likelihood estimation to empirical resampling error distributions. We obtained the latter by benchmarking the ML methods from Section~3 to 1\,000 bootstrap samples of the five regression and classification datasets, respectively. The goodness of the respective fits were determined with the \textit{Cramér-von-Mises} (CvM) criterion (see e.g., \cite{stephens}) using the \texttt{fitdistrplus} \texttt{R}-package \cite{fitdist}. One cannot expect to find a single distribution family that is always the best. On the contrary, it is to be expected that the distribution differs for different settings. However, finding a distribution family that has a good fit in many situations seems sufficient in order to build a sequential test around it. The distribution families we consider here are the normal, gamma and Weibull distributions, as well as variations of these in the form of the log-normal, log-gamma, inverse gamma and inverse Weibull distributions. %, and the inverse Weibull distribution. 

One problem that arises when using many of these distributions is that, as in the case of the logarithmic and inverse distribution families, the respective support does not contain the value 0 at all or, as in the case of the gamma distribution, has a corresponding density value of 0. This is particularly problematic in the classification context, since errors of 0 are not uncommon for certain combinations of a (sub-)dataset and a learner.
An obvious solution is to shift the data by an additive constant $c>0$. However, it must be noted that the distributions in question are generally not invariant to such shifts, i.e., the distribution family is usually not preserved. Therefore, the goodness of fit can sometimes depend strongly on the choice of $c$, especially when the observed errors tend to be small, as is the case in classification, where the errors range between 0 and 1.
To account for this, the fit of the distribution families was calculated for different shift sizes of $c \in \{0.001, 0.01, 0.1, 0.15, 0.25, 0.5, 1, 1.5\}$, and only the results for the best $c$ are reported for each distribution family. For the regression context, this problem is less relevant, since on one hand, MSE values of 0 only occur in pathological examples and on the other hand, MSE values usually have a higher magnitude compared to MMCE values and are thus less affected by small additive shifts.

\begin{figure}[h!]
\centering
\includegraphics[width=\textwidth]{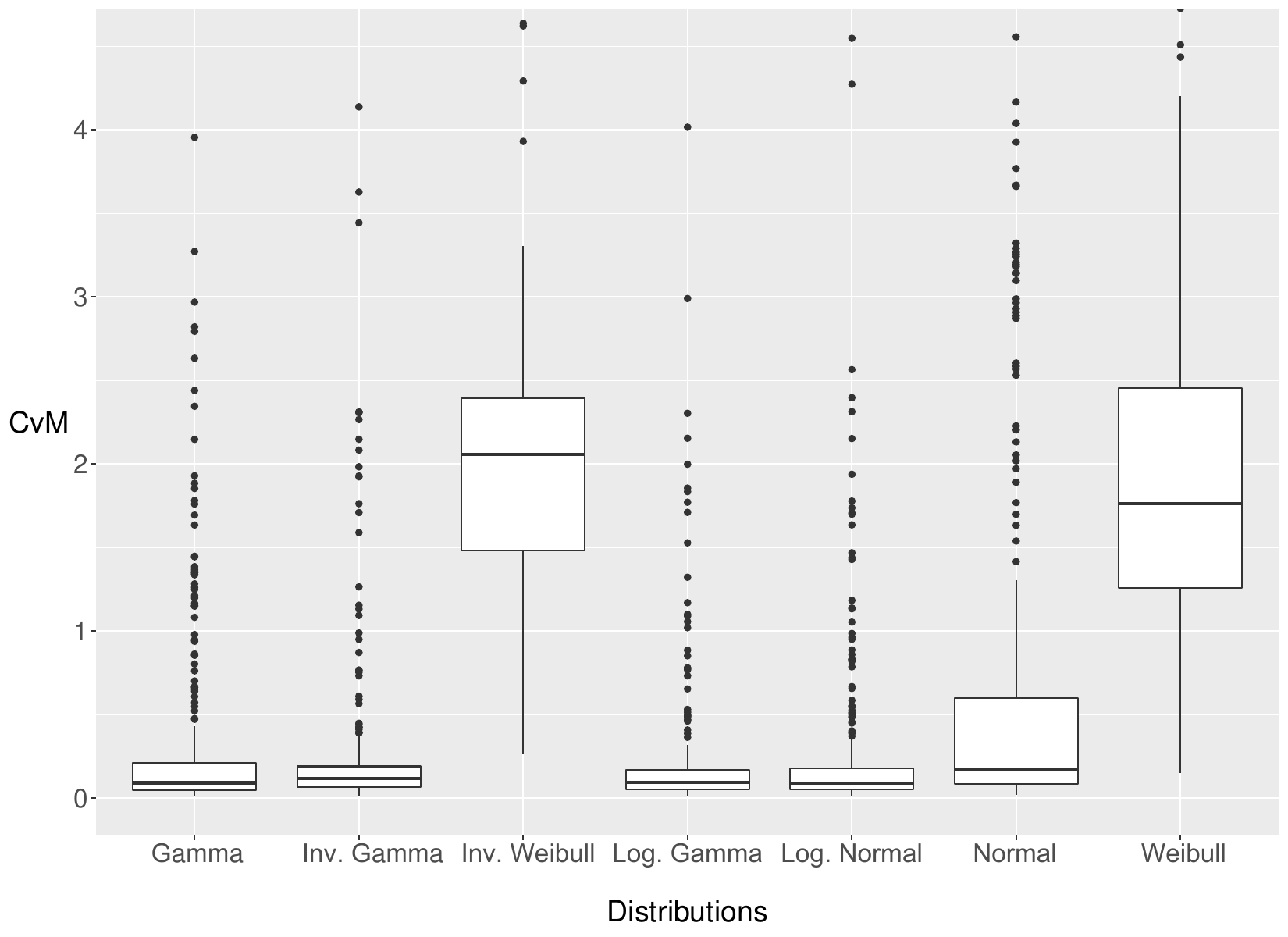}
\caption{Cramér-von-Mises criterion values for different distribution classes in the regression case.}
\label{regrAllCVM}
\end{figure}

Figure~\ref{regrAllCVM} shows the CvM values achieved by the different distribution families in the regression case. Smaller CvM values indicate a better fit. Apart from the two Weibull families and the normal distribution, a relatively homogeneous picture emerged.

\begin{figure}[h!]
\centering
\includegraphics[width=\textwidth]{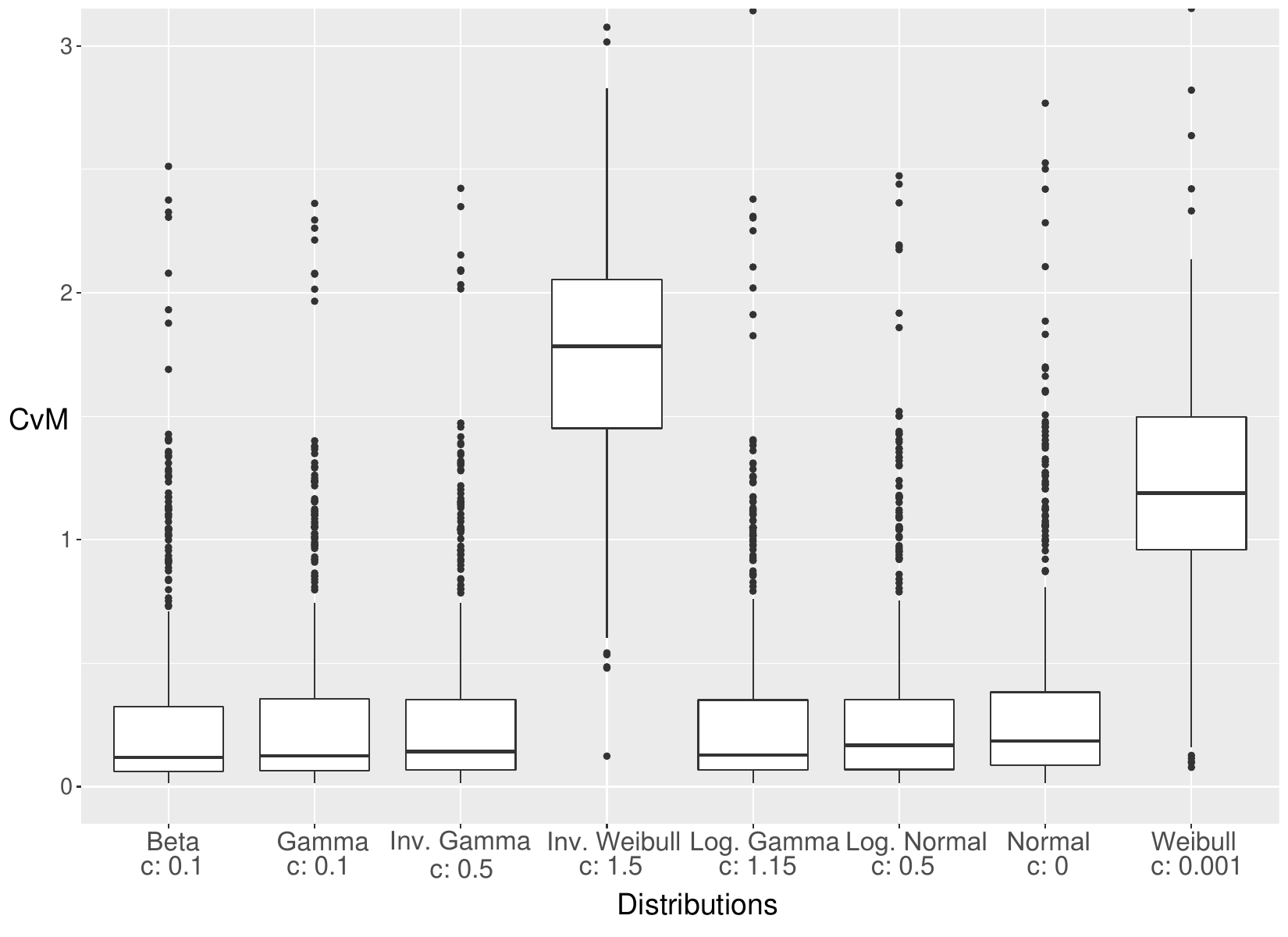}
\caption{Cramér-von-Mises criterion values for different distribution classes with individual additive shifts $c$ in the classification case.}
\label{classAllCVM}
\end{figure}

Similar results were achieved in the classification context as shown in Figure~\ref{classAllCVM}. These findings are supported when looking at other criteria such as the Kolmogorov-Smirnov and Anderson-Darling criteria (results not shown). Thus, regarding the distributional fit, the choice of the distribution family is not crucial as long as one chooses from the set of generally well-performing distribution families. 

However, out of these distribution families, the log-normal family offers two notable advantages. First, it allows performing a location test based on only one parameter since the median of a log-normally distributed variable only depends on $\mu$. Second, it allows for applying a sequential test based on a normality assumption after a logarithmic transformation of the original data. Therefore, we decided to assume a log-normal distribution for designing our sequential random search. However, due to the presence of a nuisance parameter ($\sigma^2$), regular SPRTs could not be applied. Thus, we used a \textit{Sequential Likelihood Ratio Test} (SLRT) proposed by Ghosh~\cite{ghosh} for the sequential Behrens-Fisher-problem. Here, one considers two normally distributed i.i.d. random variables $U \sim \mathcal{N}(\mu_U, \sigma_U^2)$ and $W \sim \mathcal{N}(\mu_W, \sigma_W^2)$, where all parameters are unknown and one wants to test 
\[ \text{H}_0: \gamma := \mu_U - \mu_W = \gamma_0 \ \text{vs.} \ \text{H}_1: \gamma = \gamma_1 \ \text{with} \ \gamma_0, \gamma_1 \in \mathbb{R}, \ \gamma_0 < \gamma_1,\] 
with respective type I error rate $\alpha$ and type II error rate $\beta$. The continuation region of the test is given by:
\[ - \frac{s^2_{u(n)} + s^2_{w(n)}}{\gamma_1 - \gamma_0} \ln \frac{1-\alpha}{\beta} < n \left( \bar{u}_{(n)} - \bar{w}_{(n)} - \frac{\gamma_0 + \gamma_1}{2} \right) < \frac{s^2_{u(n)} + s^2_{w(n)}}{\gamma_1 - \gamma_0} \ln \frac{1-\alpha}{\beta}, \]
where $\bar{u}_{(n)}$ and $\bar{w}_{(n)}$ denote the means and $s^2_{u(n)}$ and $s^2_{w(n)}$ the empirical variances of the respective sample including all observations up to $n$. Since it holds that
\[\mu_U - \mu_W = \gamma_0 \Leftrightarrow \frac{\exp(\mu_U)}{\exp(\mu_W)} = \exp(\gamma_0) \Leftrightarrow \frac{\text{med}(\tilde{U})}{\text{med}(\tilde{W})} = \exp(\gamma_0)\] 
for $\ln(\tilde{U}) \sim \mathcal{N}(\mu_U, \sigma_U^2)$ and $\ln(\tilde{W}) \sim \mathcal{N}(\mu_W, \sigma_W^2)$, the test above is practically a test of the relative difference between the medians of the two loss distributions (assuming log-normality) in the context of regression where no additive shift of the resampling errors is needed. In theory, one could also use a two-stage procedure instead of the SLRT, where the variances are estimated in a first step followed by the sequential test in the second step. However, this approach would require a pre-specified number of evaluations for the variance estimation alone. To keep the number of evaluation steps as small as possible, we thus opt for the SLRT. 

%TODO: Interpretation von $\gamma_0, \gamma_1, \alpha, \beta$ Reicht das?
\section{Integrating the Sequential Test into Random Search}
\label{sec:sqrs}
The general idea now is to combine the regular random search algorithm with the benefits of early stopping from the SLRT. In each iteration, a single new random hyperparameter setting is proposed and used to estimate the corresponding value of the loss function using a resampling.
However, instead of using a fixed number of resampling iterations, the resampling is continued until a statistically sound decision can be made. The new setting is compared to the current best setting until a significant difference between these two settings is found. The winner is kept as the new best setting. 
To save computational effort, evaluation results for a specific candidate are reused from previous comparisons if available. To prevent infinite run times, a maximum number of evaluations is defined. If the test cannot make a decision until this point, the setting with the smallest estimated loss function value is used. Pseudocode describing this Sequential Random Search (SQRS) procedure in detail can be found in Algorithm~\ref{sqrs}. Here, the function \texttt{generateConfig()} can be any function that generates a new hyperparameter setting. In the most simple case, uniform sampling in the search space can be used, resulting in a random search algorithms. However, it would also be possible to use more advanced optimization algorithms instead.
The function \texttt{evaluateConfig()} evaluates the performance of a single hyperparameter setting using an arbitrary resampling procedure, e.g., a single bootstrap iteration. % of bootstrapping. 
As termination criterion, we usually use a maximum number of iterations or computation time. %  should be used. 

For the implementation of \texttt{generateConfig()}, two generally different approaches can be considered: 
one possibility is to create a resampling instance with \textit{max.iter} different dataset partition samples (i.e.,\ training and test sets) at the beginning of the optimization on which all configurations are evaluated on, i.e.,\ in the $n$-th evaluation step all configurations are validated on the same bootstrap sample.
This allows immediate comparability of configurations (on the respective sample) and represents the usual approach to regular random search.
However, this approach does not do justice to the stochastic nature of the optimization problem. Therefore, the second approach is to always draw a new bootstrap sample for each evaluation of a configuration, i.e., two configurations are validated on different bootstrap samples in the $n$-th step.
In both cases, performance values already obtained for \texttt{opt.config()} are reused to save additional evaluations.

\begin{algorithm}[h!]
\caption{Sequential Random Search (SQRS)}
\label{sqrs}
\begin{algorithmic}[1]
\Procedure{SQRS}{\textit{max.iter}, $\gamma_0, \gamma_1, \alpha, \beta$}
\State \texttt{opt.config} = \texttt{generateConfig()} %\Comment{Erzeuge Lerner mit zufälligen Einstellungen.}
%\Statex \Comment{Erzeuge Lerner mit zufälligen Einstellungen innerhalb des Suchbereichs.}
%\State iter = 0
%\While {\texttt{experiments} $<$ \texttt{max.experiments}}
\While {termination condition not fulfilled}
\State \texttt{cand.config} = \texttt{generateConfig()}

\For{$n=1$ to \textit{max.iter}}
\State $p_{opt, n}$ = \texttt{evaluateConfigs(opt.config)} 
\State $p_{cand, n}$ = \texttt{evaluateConfigs(cand.config)} 
\If {$n \ge 2$}
\State \texttt{upper} = $\frac{\sigma^2(\boldsymbol{p}_{opt, 1:n}) + \sigma^2(\boldsymbol{p}_{cand, 1:n})}{\gamma_0-\gamma_1} \ln\frac{1-\beta}{\alpha}$ 
\State \texttt{lower} = $-$\texttt{upper}
\State \texttt{statistic} = $ n \cdot \left( \bar{p}_{opt, 1:n} - \bar{p}_{cand, 1:n} \right) - \frac{\gamma_0 - \gamma_1}{2}$
%\If {$\neg$ (lower $<$ statistic $<$ upper)}
\If {\texttt{statistic} $>$ \texttt{upper}} %\Comment{D.h. Annahme von $H_1$,}
\State \texttt{opt.config} = \texttt{cand.config} %\Comment{setze cand als aktuell opt. Lösung.}
\Break
\EndIf
\If {\texttt{statistic} $<$ \texttt{lower}}
\Break 
\EndIf
\If {$n = $ \textit{max.iter}} %\Comment{Wenn max. Auswertungszahl erreicht,}
\State \texttt{opt.config} =  $\argmin_{i \in \{opt, cand\}} \bar{p}_{i, 1:n}$ %\Comment{wähle Par. mit min. Mittelwert.}
\EndIf
\EndIf
\EndFor
\EndWhile 
\State \Return{\texttt{opt.config}}
\EndProcedure
\end{algorithmic}
\end{algorithm}

\section{Simulation Study}
\label{sec:sim}
We will now compare the SQRS with a regular random search in a simulation study designed to ensure maximum comparability between the two algorithms.
Using the same learners, parameter search spaces and datasets as in Tables~\ref{dataSets} and \ref{lrnTab}, we generated 50 random configurations of hyperparameters for each combination of learner and dataset, and validated their performance using a bootstrap resampling with ten iterations.
To make the resampling results comparable, we used the same resampling instance for both algorithms. We performed 100 replications. For the random search, we simply selected the parameter setting with the minimal MSE. % as its result.
For the SQRS, we did not generate new random hyperparameter settings, but instead operated on the same set of parameter configurations as for the regular random search. As a consequence, the SQRS could not find a better setting than the regular random search. 

For this simulation study, we were interested in three key aspects. First, how often does the SQRS find the optimal parameter setting? Since the SQRS is based on a statistical test with a non-zero error rate, it may erroneously discard optimal parameter settings because of an incorrect test decision due to sampling noise. Thus, it was important to investigate how often this was the case. Second, when not being able to discover the optimal setting, how much worse do parameter settings found by the SQRS perform compared to the optimal parameter settings found by random search? Third, how much faster is the SQRS compared to the regular random search?

In our simulation, we considered four SQRS settings $A$-$D$ for the regression case which varied in the choice of the SQRS parameters $\alpha,\beta \in\{0.01,\, 0.05\}$ and the location parameters $-\gamma_0 = \gamma_1\in\{0.1,\, 0.2\}$, see Table~\ref{equal.tab}. In the classification case, we considered SQRS settings $E$-$F$ where $\alpha,\beta \in\{0.01,\, 0.05\}$ and $-\gamma_0 = \gamma_1\in\{0.01,\, 0.02\}$. The choices for $\gamma_0$ and $\gamma_1$ were derived heuristically from a small simulation study that analyzed the power of the sequential test using the datasets and learners described above (results omitted here). Generally speaking, $\gamma_0$ and $\gamma_1$ are problem specific meta parameters as well. In the regression context, the choices for the location differences roughly correspond to testing whether the quotient of the two resampling error distribution medians differ by 10\% or 20\%, respectively. Note that this is not an exact relation due to the exponentiation (e.g. $e^{-0.1} = 0.904$ and $e^{0.1} = 1.105$), but it can be thought of as an intuition for the meaning of $\gamma_0$ and $\gamma_1$. Due to the additive shift needed for classification errors, $\gamma_0$ and $\gamma_1$ are not as easily interpretable as in the regression context. 

As specified by the resampling instance, the maximum number of evaluations was $\text{\textit{max.iter}} = 10$.
% We aim to answer two questions: First, can the SQRS algorithm achieve a comparable accuracy, i.e., does it find the same optimal hyperparameter settings? And second, how many functions evaluations can be saved by the SQRS compared to the regular random search?
First, we investigated how often the SQRS led to the same result as the random search. Table~\ref{equal.tab} displays the proportion of identical solutions for the different values of $\gamma_0$, $\gamma_1$, $\alpha$ and $\beta$. The SQRS chose the same parameter configuration approximately between 87\% to 90\% of the runs for the regression datasets, and between 79\% to 90\% of the time for the classification datasets. The more conservative the test became in terms of smaller (absolute) values for each test parameter, the more often the SQRS made the same decision.

%\begin{figure}
% \begin{minipage}[c]{0.4\textwidth}
\begin{table}[h!]
\center
      \caption{Proportion of identical solutions for SQRS and random search for specified $\gamma_0$, $\gamma_1$, $\alpha$ and $\beta$ values for regression (A-D) and classification (E-H).}
       \label{equal.tab}
\begin{tabular}{lccccc}
\toprule
Type & Setting & $\gamma_0$ & $\gamma_1$ & $\alpha, \beta$ & Proportion identical \\
\midrule
\multirow{4}{*}{Regression} & A & $-0.2$ & $0.2$ & $0.05$ & $0.87$ \\
& B & $-0.2$ & $0.2$ & $0.01$ & $0.89$\\
& C & $-0.1$ & $0.1$ & $0.05$ & $0.90$\\
& D & $-0.1$ & $0.1$ & $0.01$ & $0.91$\\
\midrule
\multirow{4}{*}{Classification} & E & $-0.02$ & $0.02$ & $0.05$ & $0.79$ \\
& F & $-0.02$ & $0.02$ & $0.01$ & $0.84$\\
& G & $-0.01$ & $0.01$ & $0.05$ & $0.86$\\
& H & $-0.01$ & $0.01$ & $0.01$ & $0.90$\\
\bottomrule
\end{tabular}

\end{table}

When stratifying these results by dataset and learner, Table~\ref{equal.detail} shows that the SQRS performed best at finding the identical solution for the decision tree learner. Using the most conservative settings (i.e., D for regression and H for classification), the SQRS found the optimal solution at least 94\% of the times. The results for the other learners display more variability.  In particular, the tree-based ensembles Random Forest and XGBoost are more sensitive to the settings of the SQRS. This is most notable in the classification context where choosing more conservative SQRS parameters (Setting H) greatly increased the rate at which the SQRS found the optimal solution.

\begin{table}[h!]
    \centering
        \caption{Proportion of identical solutions for SQRS and random search w.r.t. to datasets, learners and SQRS settings for regression (A, D) and classification (E, H).}
    \label{equal.detail}
    \begin{tabular}{lcccccc}
    \toprule
    \multirow{3}{*}{Dataset} & \multirow{3}{*}{Setting} & \multicolumn{4}{c}{Proportion identical}\\ \cmidrule{3-6}
&  & Decision Tree & Random Forest & XGBoost & Elastic Net\\
\midrule
\multirow{2}{*}{\textit{Boston}} & A & 0.99 & 0.81 & 0.67 & 0.75\\
& D & 0.98 & 0.84 & 0.73 & 0.79 \\[0.2cm]
%\midrule
\multirow{2}{*}{\textit{Insurance}} & A & 0.99& 0.93& 0.93 & 0.87\\ 
& D &  1.00 & 0.95 & 0.97 & 0.89\\[0.2cm]
%\midrule
\multirow{2}{*}{\textit{Diamond}}& A & 1.00 & 0.86 & 0.79 & 0.82\\
& D & 1.00 & 0.90 & 0.85 & 0.87\\[0.2cm]
%\midrule
\multirow{2}{*}{\textit{Wage}} & A & 0.88 & 0.94 & 0.91 & 0.86\\
& D & 0.94 & 0.95 & 0.94 & 0.90\\[0.2cm]
%\midrule
\multirow{2}{*}{\textit{Concrete}} & A & 0.99 & 0.91 & 0.76 & 0.76\\
& D & 0.99 & 0.95 & 0.86 & 0.79\\
\midrule
\multirow{2}{*}{\textit{German Credit}} & E & 0.90 & 0.72 & 0.64 & 0.78\\
& H & 0.96 & 0.87 & 0.81 & 0.91\\[0.2cm]
%\midrule
\multirow{2}{*}{\textit{Phoneme}} & E &  1.00 & 0.63 & 0.67 & 0.92\\
& H & 1.00 & 0.79 & 0.88 & 0.95\\[0.2cm]
%\midrule
\multirow{2}{*}{\textit{Pima Indians}} & E &  0.84 & 0.75 &  0.80 & 0.84\\
& H & 0.94 & 0.84 & 0.92& 0.88\\[0.2cm]
%\midrule
\multirow{2}{*}{\textit{Cancer}} & E & 0.91 & 0.71 & 0.68 & 0.74\\
& H & 0.97 & 0.86 &  0.85 & 0.90\\[0.2cm]
%\midrule
\multirow{2}{*}{\textit{Ionosphere}} & E & 0.96 & 0.74 & 0.73 & 0.82\\
& H & 0.98 &  0.88 &  0.86 & 0.85\\
    \bottomrule
    \end{tabular}

\end{table}

%   \begin{minipage}[c]{0.55\textwidth}
%     \centering
% \includegraphics[scale = 0.33]{plots/perf_ratio_regr_ABCD.pdf}
% \captionof{figure}{Performance ratio of SQRS/random search in the case of different solutions (i.e., performance ratio SQRS/random search $\neq$ 1) for the four different settings.}
% \label{perf.ratios}
%     \end{minipage}
% \end{figure}

\begin{figure}[h]
\centering
\includegraphics[width=\textwidth]{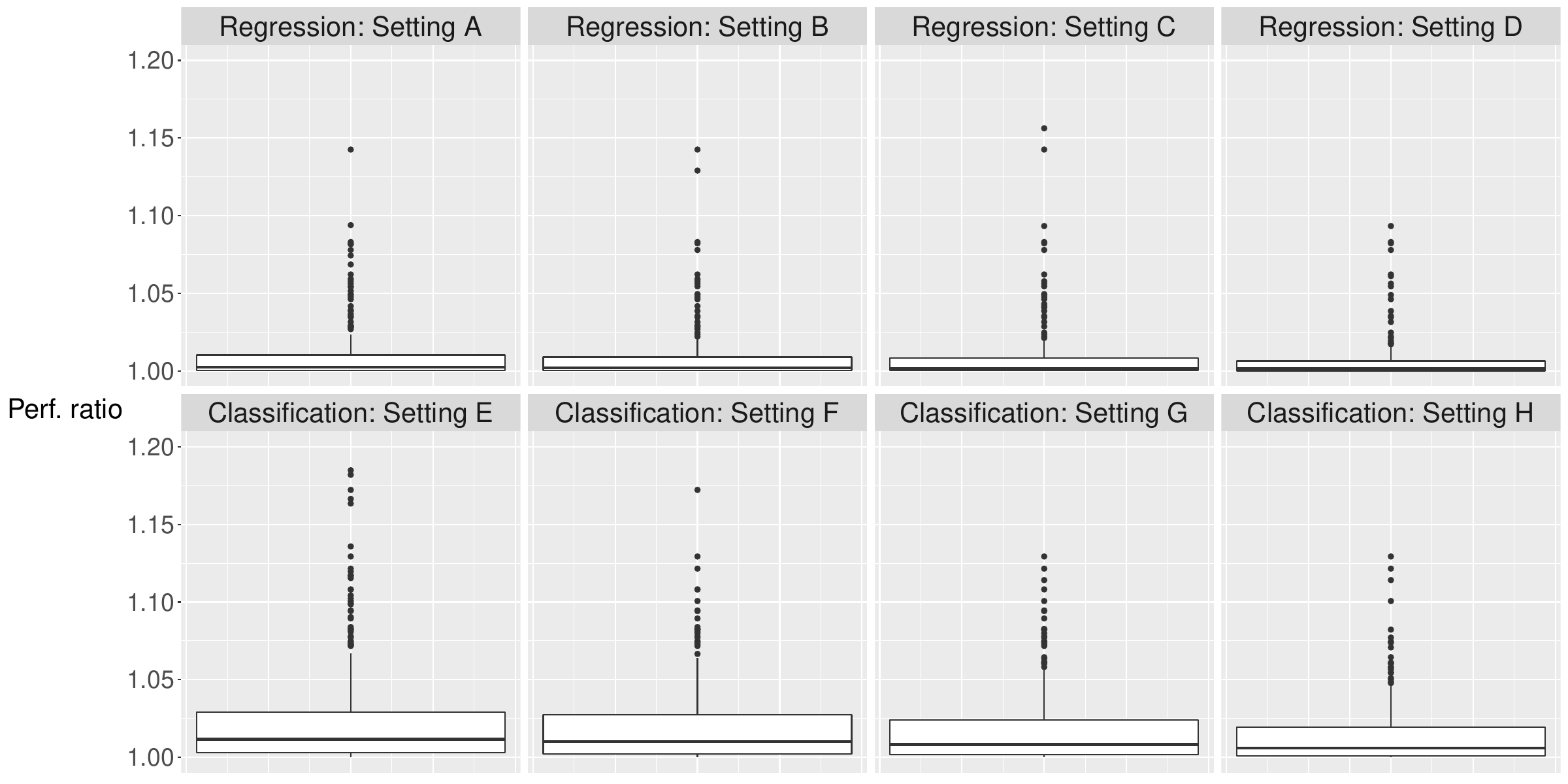}
\caption{Performance ratios of SQRS/random search in the case of different solutions (i.e., performance ratio SQRS/random search $\neq$ 1) for regression (A-D) and classification (E-H) SQRS settings.}
\label{perf.ratios}
\end{figure}

If we consider the cases in which the SQRS did not find the "optimal" solution (i.e.\ the solution obtained by the random search), the solutions found by the SQRS did not perform considerably worse.
Figure~\ref{perf.ratios} shows the quotient of the SQRS and random search performances.
A ratio of 1 corresponds to identical performances, ratios larger than 1 indicate a better performance of the random search. Per definition of our experiment, SQRS could not find a better setting than the random search, hence, ratios smaller than 1 were not possible here. For better clarity, we omit all data points with a ratio of 1 in the plot. It can be seen that in cases where SQRS and random search differed, the solution found by SQRS is only marginally worse, especially in the context of regression.

\begin{figure}[h]
\centering
\includegraphics[width=\textwidth]{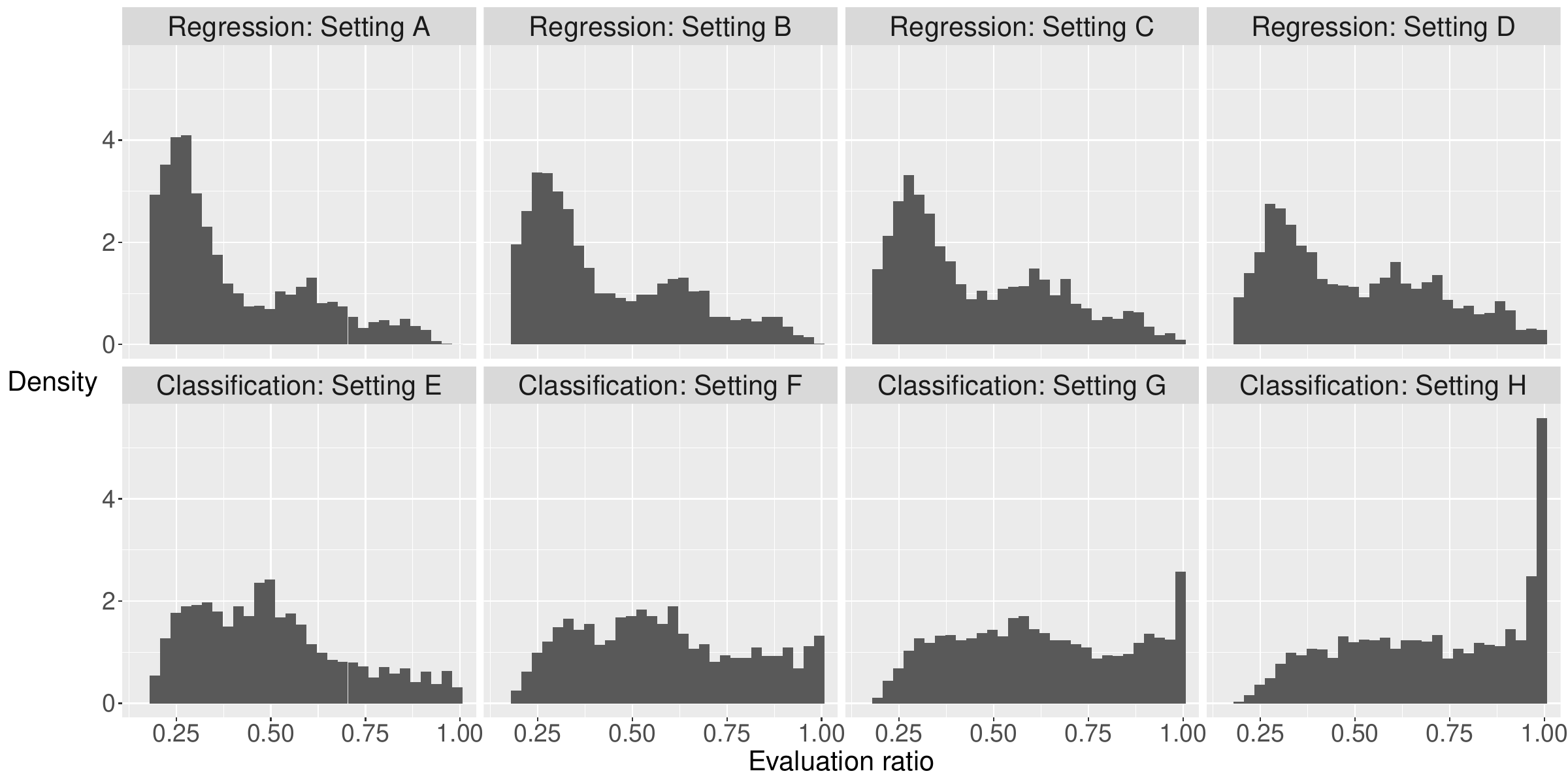}
\caption{Ratios of evaluations needed for SQRS/random search for regression (A-D) and classification (E-H) SQRS settings.}
\label{evalRatio}
\end{figure}

Figure~\ref{evalRatio} shows the number of evaluations SQRS needed to reach its top performance compared to the total number of random search iterations. The quotient of the required evaluations for the sequential and the regular random search is (at the median) 0.32 for Setting A, 0.36 for Setting B, 0.39 for Setting C, and 0.46 for Setting D. In all scenarios, the SQRS never exhausted the maximum allowed number of evaluations (i.e., it did not reach an evaluation ratio of 1). For the liberal setting (A), at most half as many evaluations were needed as compared to the regular random search in 71.1\% of the runs. For the conservative setting (D), this was the case in 55.3\% of all runs.

For classification problems, generally more evaluations were needed. For Setting E, we achieved a median evaluation ratio of 0.48, for Setting F 0.56, for Setting G 0.61 and for Setting H 0.71. The percentage of runs in which the maximum number of evaluations was exhausted, ranged between 0.15\% (Setting E) and 9.2\% (Setting H).

% In the case of classification, Figure~\ref{evalRatioClassif} shows that 
% \begin{figure}[h]
% \centering
% \includegraphics[width=\textwidth]{plots/classif_eval_hists.pdf}
% \caption{Ratios of evaluations needed for SQRS/random search in the classification context.}
% \label{evalRatioClassif}
% \end{figure}

\newpage
\section{Discussion}
\label{sec:dis}
In this work, we analyzed the feasibility of employing sequential statistical tests during the hyperparameter tuning process to save computational effort. We aimed at answering two main research questions. The first one pertained to the construction of a suitable sequential test for hyperparameter tuning. To study what kind of approximate parametric assumption could be made for the resampling error distributions, we performed a small simulation study in which we fitted multiple different distribution families to empirical resampling error distributions. Overall, typical flexible distribution families achieved comparably good fits: the gamma, the inverse gamma, the log-gamma and the log-normal distribution. We recognize that this approach was purely empirical, but it provided us with sufficient results to continue our work upon.  % and requires a more thorough analysis in future but it provided us with a first indication to continue our work upon. 

Although multiple distribution families appeared suitable, we decided for the log-normal family for practical reasons. Using a sequential test by \cite{ghosh}, we implemented a sequential variant of the random search (abbreviated SQRS). The main difference between the SQRS and regular random search was the early stopping possibility of the former. After each evaluation step, the resampling errors samples of two configurations were compared using the sequential test procedure. When a terminating decision could not be made before reaching the maximum number of permitted evaluation steps, the procedure selected the parameter setting leading to the smaller resampling error. 

Having implemented the SQRS, our focus then turned to the second research question of how a hyperparameter tuning approach using a sequential statistical test would perform compared to a regular random search. We performed a simulation experiment in which the SQRS and regular random search tackled the same regression and classification problems under identical conditions. We found that by using the sequential testing procedure instead of a full resampling, the number of evaluations could be greatly reduced without considerable performance loss. For the regression problems we studied, we could cut the needed evaluation steps by more than half, for classification problems we could save between 29\% and 53\% of evaluations.   

We recognize our experiments are just a first proof of concept. Our comparisons were aimed at maximal comparability since both algorithms were forced to operate under the same laboratory conditions using an identical set of candidate settings. We are fully aware that our SQRS algorithm loses the most important advantage of random search: the ability to use (nearly) unlimited parallel computation power. Nonetheless, we believe that our results indicate that the inclusion of a sequential statistical test procedure within a hyperparameter tuning algorithm is a promising approach and we see many opportunities for further research. 

For example, since every tuning algorithm has to compare hyperparameter settings at its core in order to decide for the best one, the sequential test could be integrated into many existing tuning algorithms. It would be interesting to study whether replacing the non-sequential procedures that are typically used with our proposed sequential test yields runtime advantages. 

\begin{figure}[h!]
    \centering
    \includegraphics[width=\textwidth, trim=4cm 18.5cm 1.5cm 3cm]{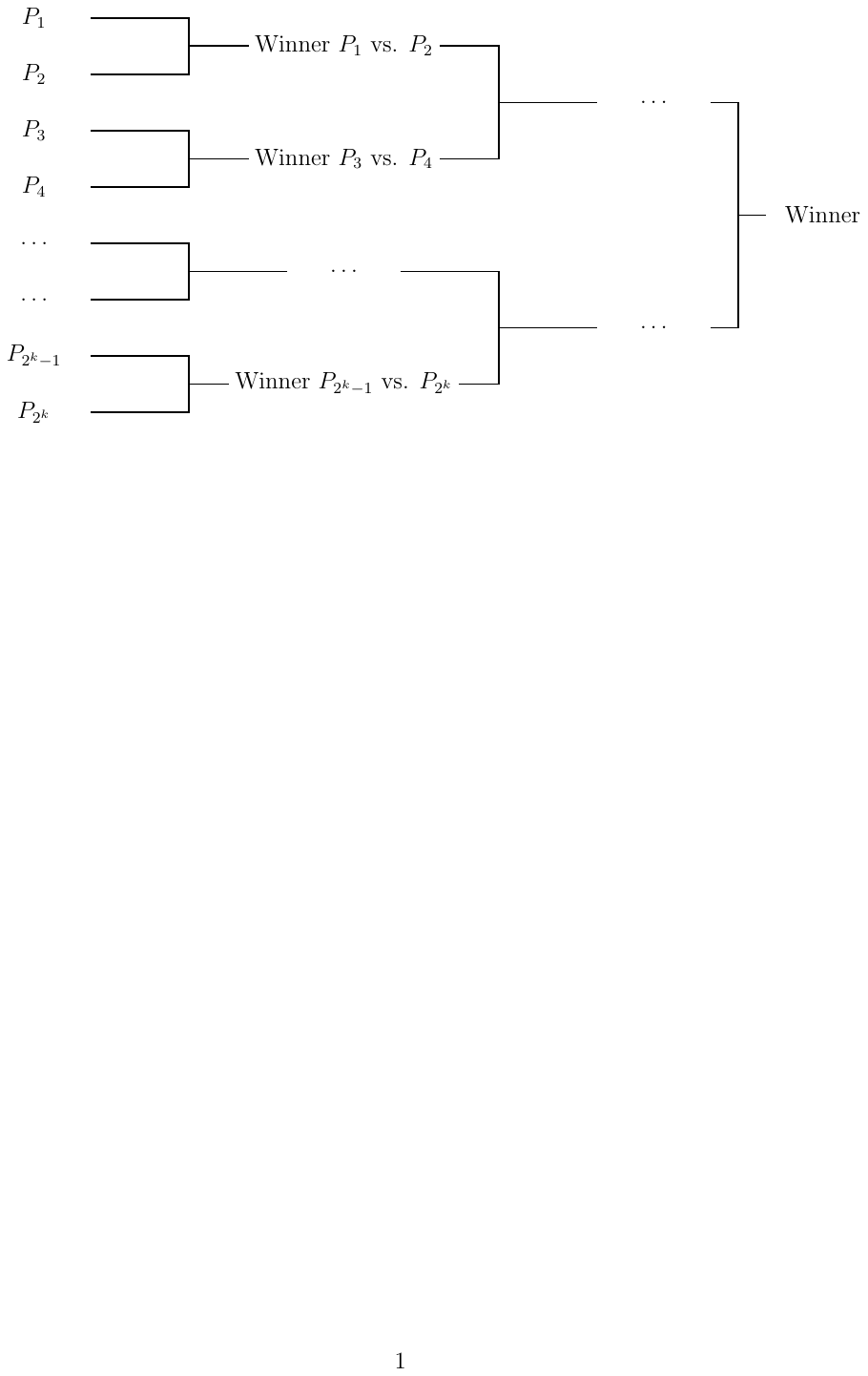}
    \caption{Exemplary parallel variant of the SQRS for parameter configurations $P_1, \dots, P_{2^k}$.}
    \label{bracket}
\end{figure}

Further, the sequential test procedure also lends itself to be used for parallel computation. One possibility could be to implement a procedure resembling a bracket from knockout tournaments in sports competitions as depicted in Figure~\ref{bracket}. For $2^k$ different parameter configurations $P_1, \dots, P_{2^k}$ to be tested, the individual duels could be executed in a parallel fashion. Of course, the level of parallelization is less than for a default random search. For other algorithms, individual solutions would have to be developed. We believe that for most algorithms the sequential testing procedure can be used while still allowing for parallel computing.

We also think it is important to consider the reduced computational effort from the perspective of sustainability. Solving a hyperparameter tuning problem using a highly parallelized random search may be a simple and efficient approach. However, it is also an approach that consumes a lot of resources (i.e., electrical power). More economical approaches are reasonable, and we hope that our approach can contribute here. Overall, we believe that there remains a lot of untapped potential in integrating sequential test procedures into hyperparameter tuning that warrants further investigation in future work.

\backmatter

\bibliography{sn-bibliography}% common bib file


\end{document}